%% file: paper_full.tex
\documentclass{article}

\usepackage[preprint]{corl_2022} 

\usepackage{subcaption}
\usepackage{graphicx}
\usepackage{algorithm}
\usepackage[noend]{algpseudocode}
\usepackage{amsmath}
\usepackage{amsfonts}
\usepackage{hyperref}
\usepackage[nolist]{acronym}
\usepackage[toc,page]{appendix}
\usepackage{pifont}
\usepackage{hhline}

\title{Dext-Gen: Dexterous Grasping in Sparse Reward Environments with Full Orientation Control}

\author{
  Martin Schuck\thanks{These authors contributed equally to this work}\\
  Technical University Munich \\
  \texttt{martin.schuck@tum.de} \\
  \And
  Jan Br\"udigam\footnotemark[1]\\
  Technical University Munich \\
  \texttt{jan.bruedigam@tum.de} \\
  \AND
  Alexandre Capone\\
  Technical University Munich \\
  \texttt{alexandre.capone@tum.de} \\
  \And
  Stefan Sosnowski\\
  Technical University Munich \\
  \texttt{sosnowski@tum.de} \\
  \And
  Sandra Hirche\\
  Technical University Munich \\
  \texttt{hirche@tum.de} \\
  }

\begin{document}
\maketitle
\input{acronyms.tex}


\begin{abstract} 
    Reinforcement learning is a promising method for robotic grasping as it can learn effective reaching and grasping policies in difficult scenarios. However, achieving human-like manipulation capabilities with sophisticated robotic hands is challenging because of the problem's high dimensionality. Although remedies such as reward shaping or expert demonstrations can be employed to overcome this issue, they often lead to oversimplified and biased policies. 
    We present Dext-Gen, a reinforcement learning framework for Dexterous Grasping in sparse reward ENvironments that is applicable to a variety of grippers and learns unbiased and intricate policies. Full orientation control of the gripper and object is achieved through smooth orientation representation. Our approach has reasonable training durations and provides the option to include desired prior knowledge. The effectiveness and adaptability of the framework to different scenarios is demonstrated in simulated experiments.
\end{abstract}



\section{Introduction}
Besides the stationary industrial setting, robots have not been widely used for object grasping and manipulation. While advances in learning and control have lead to significant progress in navigation tasks---such as Boston Dynamics' robots traversing difficult terrain \cite{Griffin2019}---robotic grasping capabilities are still underdeveloped, especially when it comes to dexterous manipulators like anthropomorphic robotic hands. As a result, robots are severely limited in real-world applications. Nevertheless, deep reinforcement learning is a promising method for obtaining such complex grasp policies \cite{Kleeberger2020}. However, due to the high dimensionality and complexity of the grasping problem, developing effective reinforcement learning methods remains challenging.

\paragraph{Related work}

Approaches for grasping based on reinforcement learning are interesting due to their inherent closed-loop nature and nuanced grasping strategies. At the same time, a fundamental problem in applying reinforcement learning to grasping is that grasp success is extremely unlikely with initially random behavior, yet it is necessary for the agent to learn successful policies. 

A common strategy to overcome this issue is to include prior knowledge about grasping in the reward to increase the success probability, for example by shaping a reward function to promote stable grasps. This principle is used in simulation with object pose information and trust-region policy optimization \cite{Merzic2019}, or with proximal policy optimization \cite{Schulman2017, Wu2019, Shahid2020} to learn grasping from complex reward functions. While reward shaping is helpful to guide training, it introduces a bias in the training towards policies that are effective at maximizing the specific reward function, but not necessarily optimal for accomplishing the desired task \cite{Rajeswaran2018, Trott2019}. As the complexity of a robot's environment grows, it becomes increasingly difficult to find a reasonable reward shape that matches the intended task and provides sufficient guidance during training.

Another idea for facilitating initial learning is introducing off-policy expert knowledge. Grasping and other manipulation tasks can then be trained with natural policy gradients augmented with a penalty term for deviations from human demonstrations previously recorded in virtual reality with motion capture \cite{Rajeswaran2018}. Similarly, an expert dataset can be generated from standard motion planning and grasp planning tools to penalize policies that differ from the expert dataset and thereby guide the agent's exploration \cite{Wang2021}. Along the same line of reasoning, teleoperation trajectories can be recorded to improve the initial grasping guess with path integral policy improvement \cite{montforte2021}. A drawback of these approaches is that expert knowledge must be obtained for each task and imposes a data-driven prior on the training process. This prior introduces a bias towards policies that are close to the demonstrations, but might not be optimal for the agent's task.

Instead of relying on external insight into complex problems, hindsight learning improves the sample efficiency of sparse reward experience. The intended goal is swapped with an achieved goal during training to obtain samples with a positive reward \cite{Andrychowicz2017, Plappert2018}. While this concept is evaluated in simulation and the transfer to real systems is demonstrated, the algorithms and examples in \cite{Andrychowicz2017} related to object relocation are limited to a parallel jaw gripper with fixed orientation and disregard the object's goal orientation. 

For a more detailed treatment of the related grasping literature, see Appendix A.

\begin{figure}[t!]
    \centering
    \begin{subfigure}[b]{0.45\linewidth}
        \centering
        \includegraphics[width=.7\linewidth]{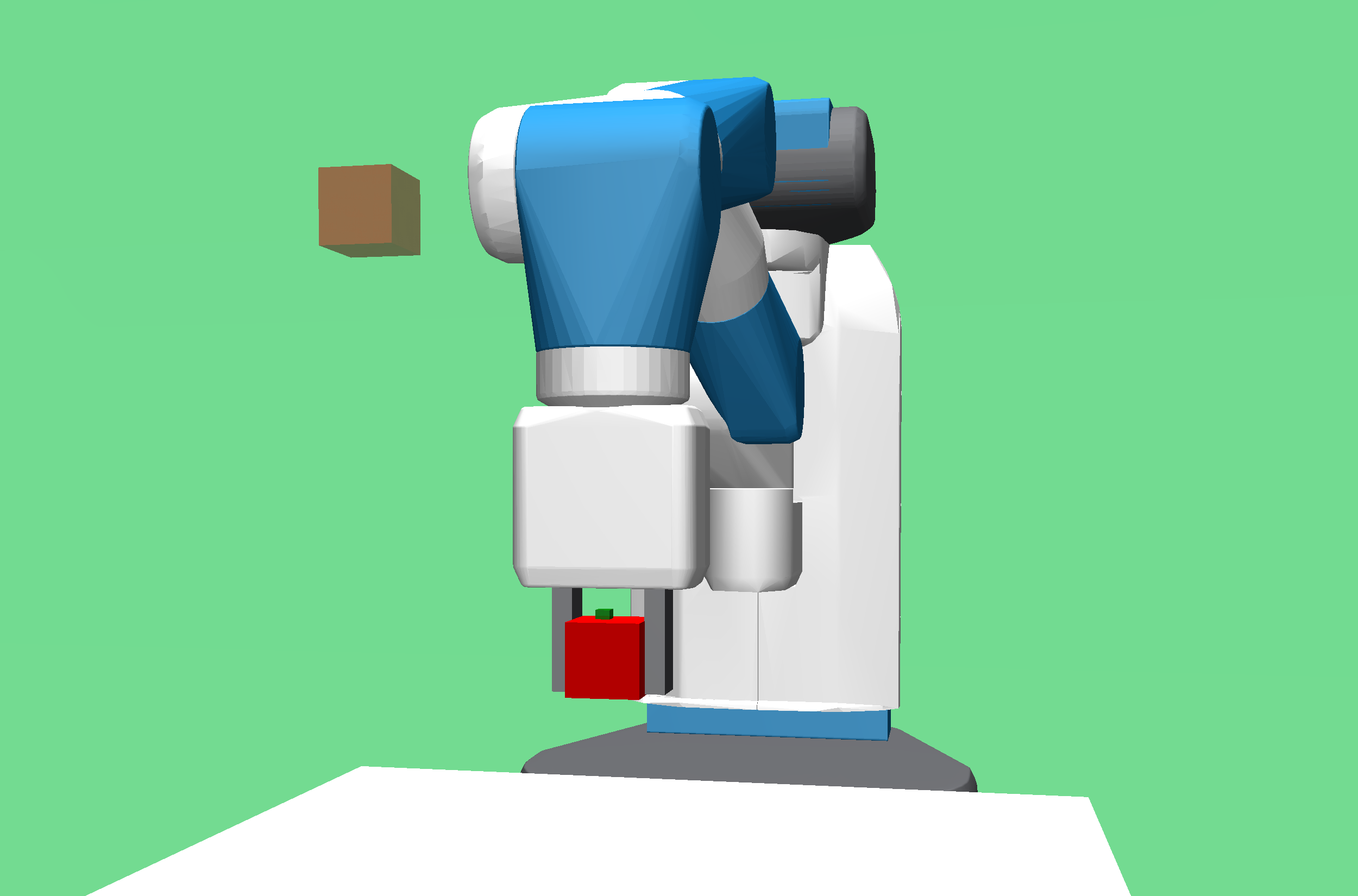}
        \label{fig:a}
    \end{subfigure}\hfill
    \begin{subfigure}[b]{0.45\linewidth}
        \centering
        \includegraphics[width=.7\linewidth]{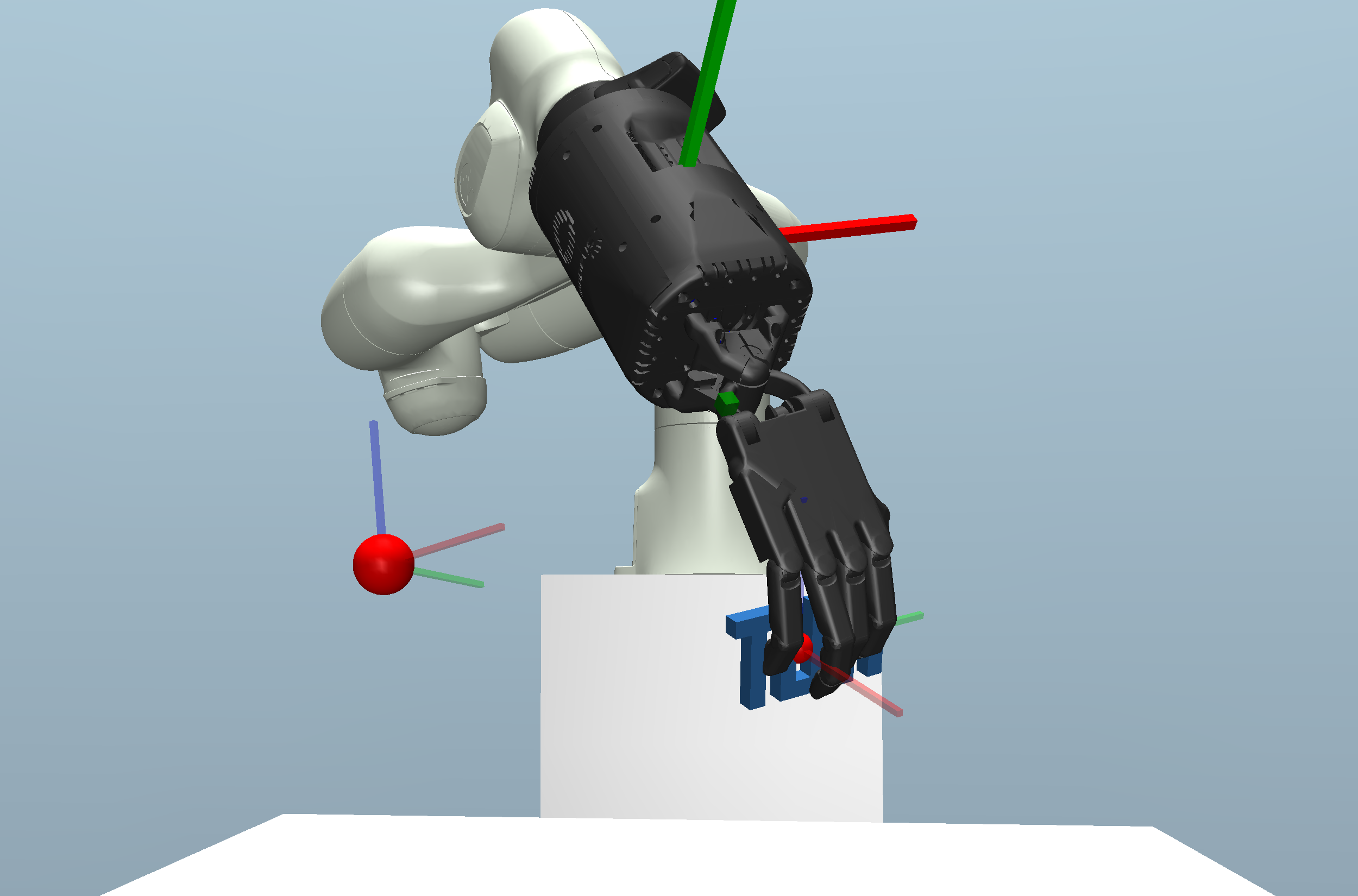}
        \label{fig:b}    
    \end{subfigure} 
    \caption{Comparison of the simple grasping task from \cite{Andrychowicz2017} with grasping based on our learning framework. (left) Pick and place with fixed-orientation for both the parallel jaw gripper and the object start pose. (right) Complex grasping task with an anthropomorphic robotic hand, full orientation control and random start poses.}
    \label{fig:comparison}
\end{figure}

\paragraph{Contribution} 
Current state-of-the-art methods for grasping either require specific reward functions, expert knowledge, or are limited to simple tasks, i.e., they disregard the gripper's and object's goal orientation. Consequently, we build on the concept of hindsight learning and extend it to learning six-dimensional grasp poses and object goal orientations with complex grippers. The result is a comprehensive grasping framework that incorporates many desireable features for learning robotic grasping. A comparison to current state-of-the-art approaches is shown in Table \ref{tab:comparison}, and an exemplary visualization of the difference in task complexity is displayed in Fig. \ref{fig:comparison}. Our framework can be used as a basis for learning a broad class of real-world robotic grasping tasks and takes a significant step towards improving robotic manipulation capabilities.

\begin{table}[h]
    \begin{center}
        \setlength\tabcolsep{5pt}
        \begin{tabular}{l || c c c | c} 
        & \citet{Rajeswaran2018} & \citet{Andrychowicz2017} & \citet{Wu2019} & Ours \\
        \hhline{=||===|=}
        Gripper type & \textbf{dexterous} & parallel jaw & \textbf{dexterous} & \textbf{dexterous} \\
        Palm orientation & \textbf{free} & fixed & 1 DoF & \textbf{free} \\ 
        Goal orientation & \textbf{controlled} & ignored & ignored & \textbf{controlled} \\ 
        Prior knowledge & required & not used & required & \textbf{optional} \\
        Reward function & \textbf{sparse} & \textbf{sparse} & shaped & \textbf{sparse} \\
        \end{tabular}
    \end{center}
    \caption{Comparison to grasping approaches in reinforcement learning. Desired properties bold.}
    \label{tab:comparison}
\end{table}

\paragraph*{Structure} The rest of the paper is structured as follows: We present our grasping framework in Sec. \ref{sec:pipeline}. Subsequently, the framework is evaluated on several grasping tasks in Sec. \ref{sec:results}. Limitations and conclusions are stated in Sec. \ref{sec:limitations} and Sec. \ref{sec:conclusion}, respectively.

\section{Learning-based dexterous grasping framework} \label{sec:pipeline}
\begin{figure}[H]
	\centering
	\includegraphics[width=\linewidth]{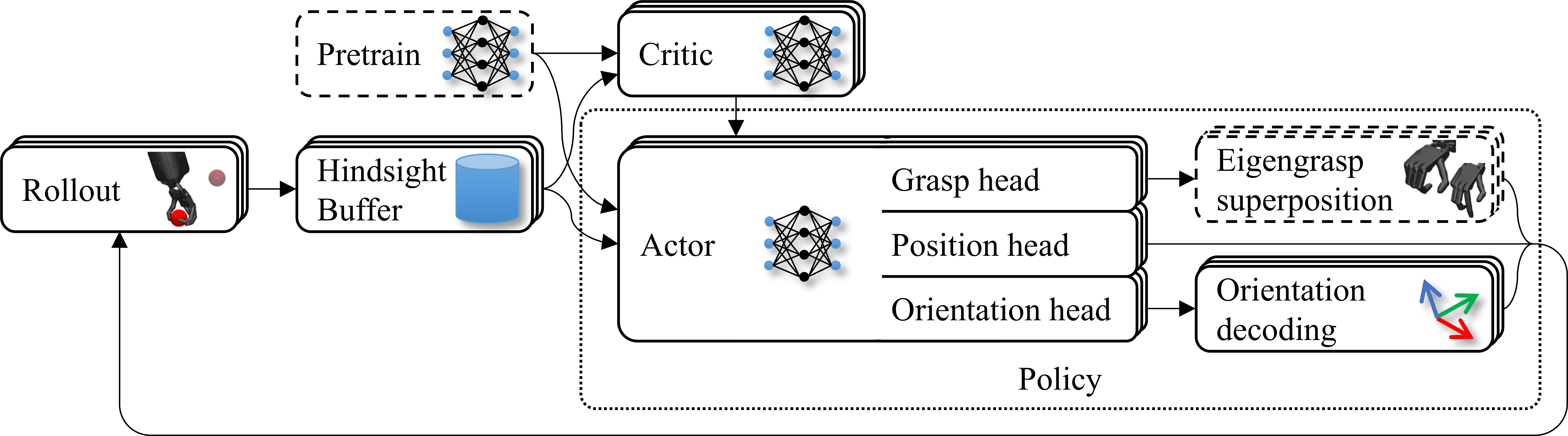}
	\caption{The grasping framework. Distributed rollout samples are collected in hindsight learning replay buffers which encode the raw states. The actor and critic networks can optionally be initialized with a previously trained network or trained from scratch. The dimension of the grasp head (output) of the actor network can optionally be reduced by a superposition of eigengrasps. The six-dimensional orientation representation from the orientation head is decoded into a proper orientation. The resulting policy is used for the next rollouts.}
	\label{fig:pipeline}
\end{figure}

We present a comprehensive reinforcement learning framework for grasping that is applicable to both simple and sophisticated grippers. The sparse reward formulation and full orientation control allow for effective grasping policies unbiased by a reward shape or demonstrations. A combination of hindsight learning, the option to include desired prior knowledge through eigengrasps and pre-training, and the distributed learning structure results in reasonable training times and convergence rates. The overall architecture is depicted in Fig. \ref{fig:pipeline}.  This section briefly introduces the basic theoretical concepts of our learning setup, and then describes the individual framework components with their respective significance in detail.


\paragraph{Multi-goal Markov decision processes}
The grasping task is formulated as a \ac{MDP}, a stochastic control process where an action $a$ is chosen at each time step according to a policy $\pi$. \acp{MDP} are formally defined by their states $s\in\mathcal{S}$, actions $a\in\mathcal{A}$, a reward function $r: \mathcal{S} \times \mathcal{A} \rightarrow \mathbb{R}$, the state transition dynamics $P: \mathcal{S} \times \mathcal{A} \times \mathcal{S} \rightarrow \mathbb{R}$, and the initial state distribution $\rho_0: \mathcal{S} \rightarrow \mathbb{R}$. The objective is to learn the optimal policy $\pi^*: \mathcal{S} \rightarrow \mathcal{A}$ that maps observed states to actions and maximizes the agent's expected future reward $\mathbb{E}\left[\sum_{t=0}^\infty \gamma^t r\left(s_t,a_t\right)\right]$, with a discount factor $\gamma\in[0, 1)$.

We leverage hindsight experience for learning in sparse reward environments by extending the \ac{MDP} to include an episode-specific goal $g \in \mathcal{G}$, as explained by \cite{Andrychowicz2017}. The multi-goal reward $r_g: \mathcal{S} \times \mathcal{A} \times \mathcal{G} \rightarrow \mathbb{R}$ then depends on the current goal and the agent needs to learn the goal-specific optimal policy $\pi^*_g: \mathcal{S} \times \mathcal{G} \rightarrow \mathcal{A}$. The reward in our case is 0 if the agent has managed to lift its target object within a distance of 0.05m to the goal position, and -1 otherwise.


\paragraph{Deep deterministic policy gradients with hindsight learning}
Precise control in complex state spaces as encountered in robotic grasping requires expressive policies with continuous action values. In this paper we utilize \ac{DDPG}, a deep reinforcement learning, actor-critic algorithm for continuous state and action spaces \cite{Lillicrap2016}.

\ac{DDPG} learns a deterministic actor policy network $\pi_\theta(s)$ and a critic network $Q_\phi(s, a)$ that approximates the Q-function, where $\theta$ and $\phi$ are the network parameters.
These parameters are trained with the deterministic policy gradient algorithm \cite{Silver2014} and temporal difference learning \cite{SuttonB1998}, respectively.

\ac{DDPG} can be extended to learn goal-specific policies by using universal value function approximators that include the current goal in the network inputs, so that a goal-specific policy $\pi_{g,\theta}(s, g)$ and critic $Q_{g,\phi}(s, g, a)$ are obtained \cite{Schaul2015}. During training, the networks are optimized with samples $e_t(s_t, g_t, a_t, r_t, s_{t+1})$ from a replay buffer $\mathcal{D} = \{e_1, ..., e_n\}$. However, because of the large dimensionality of the state and goal space $\mathcal{S} \times \mathcal{G}$ and the sparsity of the reward function $r_g$, it is highly unlikely that the agent achieves a successful grasp at random. Hindsight learning alleviates this issue by additionally storing the achieved goal in the replay buffer \cite{Andrychowicz2017}. When a batch is sampled, a part of the desired goals (around 80\%) is replaced with goals that the agent actually achieved during the episode. These virtual goals generate sufficient non-negative rewards even in sparse reward environments to enable learning.

\paragraph{Six-dimensional gripper palm poses} \label{subsec:6dof}
While grasping has been extensively studied, many approaches are limited to top down grasps or orientations on a plane to circumvent dealing with a six-dimensional palm pose of the gripper \cite{mahler2017, Wu2019, Merzic2019}. Additionally, the choice of orientation representation is crucial for learning success, as the smoothness of this representation impacts how well a network is able to approximate it \cite{Xu2004}. Our policy network learns an orientation representation that smoothly varies in $SO(3)$.

Generally, some commonly used orientation representations in robotics, such as Euler angles or quaternions, are discontinuous, in the sense that the desired network output exhibits discontinuities even for only a small change in the input at certain configurations\footnote{Euler angles are discontinuous at angles of $\mod\left(\frac{\pi}{2}\right)$. Quaternions are not unique due to their double cover of $SO(3)$.}. Therefore, such representations are not well suited for neural networks \cite{Zhou2019}.

To avoid the aforementioned drawbacks, we propose a novel actor network architecture that uses three action heads and preserves the full expressiveness of general orientations in $SO(3)$. The first and second head output the gripper joint angles and the desired cartesian translation of the gripper pose, respectively.
The third head uses the continuous six-dimensional orientation representation from \cite{Zhou2019} and recovers a full rotation matrix from this representation. To account for network saturation and singular representation results, an additional regularization term is used for training. A detailed description of the orientation head implementation is given in Appendix C.

\paragraph{Reducing the action dimensionality with eigengrasps} \label{subsec:eigengrasps}
Dexterous robotic gripper policies are especially challenging to learn because of their high dimensionality, with some anthropomorphic robotic hands featuring up to 20 actuated joints \cite{shadowhand}. However, it has been observed in humans that the majority of grasps can be explained by a lower-dimensional subspace of the hand's configuration space \cite{Santello1998, Bicchi2011, Ficuciello2014}.
We allow for the inclusion of such information in our framework to reduce the agent's action space dimensionality for complex grippers in case such prior knowledge is desireable for a given task.

Eigengrasps (also called postural hand synergies) are obtained by extracting the principal components from a dataset of successful grasp examples by principal component analysis \cite{Ficuciello2014}. A superposition of the first $N$ eigengrasps can be sufficient to explain most grasps of the human hand, with the first two principal components explaining about 80\% of the data variance \cite{Santello1998}. Using these principal components, the agent executes a weighted superposition of hand synergies to control the gripper. The synergies also conveniently enable us to encode prior knowledge about common grasp shapes and finger combinations into the agent's action space.

Eigengrasps for grippers can be manually designed or obtained from datasets. In this paper, we use the recently published ContactPose dataset \cite{Brahmbhatt2020} to match robotic hand configurations to human grasp poses. Due to the kinematic similarity between anthropomorphic robotic hands and human hands, we assume that imitated human grasps are also suitable poses for robotic hands. These eigengrasps can then be used as the gripper action space of the agent. For further details on extracting eigengrasps, see Appendix C.

\paragraph{Transfer learning with pre-training}
During regular training runs, the agent learns to grasp a specific object from scratch. To speed up training times, we can reuse networks that were previously trained on generic objects, for example a ball, for the network initialization. Similar concepts have been explored under the wider framework of transfer learning in reinforcement learning \cite{Zhu2020, Lazaric2012}.

The reutilization of networks is based on the assumption that the shape of the universal value function and the optimal policy are close to each other in most areas of the state and goal space $\mathcal{S} \times \mathcal{G}$.
In our view, this assumption holds since the general grasping strategy of approaching the object, grasping it and moving to the  goal is independent of the exact object shape. Furthermore, pre-training promotes convergence because the exploration is based on a strategy that already incorporates the general idea of grasping and thereby generates more useful training samples than a random policy.

\paragraph{Distributed training} \label{subsec:distributed}
While hindsight learning significantly increases the training sample efficiency, we still require a number of samples in the order of $10^7$ to train our agents. In order to simulate this much training data within a reasonable time frame, we use distributed reinforcement learning on multiple training nodes with PyTorch \cite{Paszke2019} and OpenMPI \cite{Graham2005}.
Our implementation is based on the works of \cite{baselines2017} and the OpenAI gym environments from \cite{Plappert2018}.

We run 16 training nodes, each of which samples multiple trajectories and appends the experience to a local experience replay buffer. During network optimization, the nodes draw a batch from their own buffer, modify the goals and recalculate the rewards to generate hindsight experience.
After each network optimizer pass, we then synchronize the gradients across all nodes so that the networks stay synchronized. The initial networks are created on the first node and broadcast to the rest to start with the same weights on all nodes. Since we average the gradients, the effective batch size during training is multiplied by a factor of 16.
We also include distributed normalizers for the states to address network sensitivities to input data scaling. The exact algorithm and hyperparameters can be found in Appendix C.

\section{Experimental results} \label{sec:results}
To demonstrate the general applicability to different grippers, we test our grasping framework in several environments using three end effectors: a parallel jaw gripper, the BarrettHand, and the ShadowHand.
All environments are implemented in MuJoCo, a high performance simulator for research in robotics with realistic contact physics~\cite{Todorov2012}. During the beginning of an episode, a target object is placed on a table with random orientation at a random position, and a random goal pose is sampled either on or above the table with equal probability. We use balls, cylinders, cubes, and an object with a more complex shape to demonstrate that the approach is able to deal with different object geometries. Since the agent only gets pose information about the target object, it still has to learn each object separately. We average the results from 5 runs with different random seeds for each experiment. Success rates are averaged across 160 test runs with the greedy policy. For details on training hyperparameters, see Appendix C.

\begin{figure}[ht!]
    \centering
    \begin{subfigure}[b]{0.2\linewidth}
        \centering
        \includegraphics[width=\linewidth]{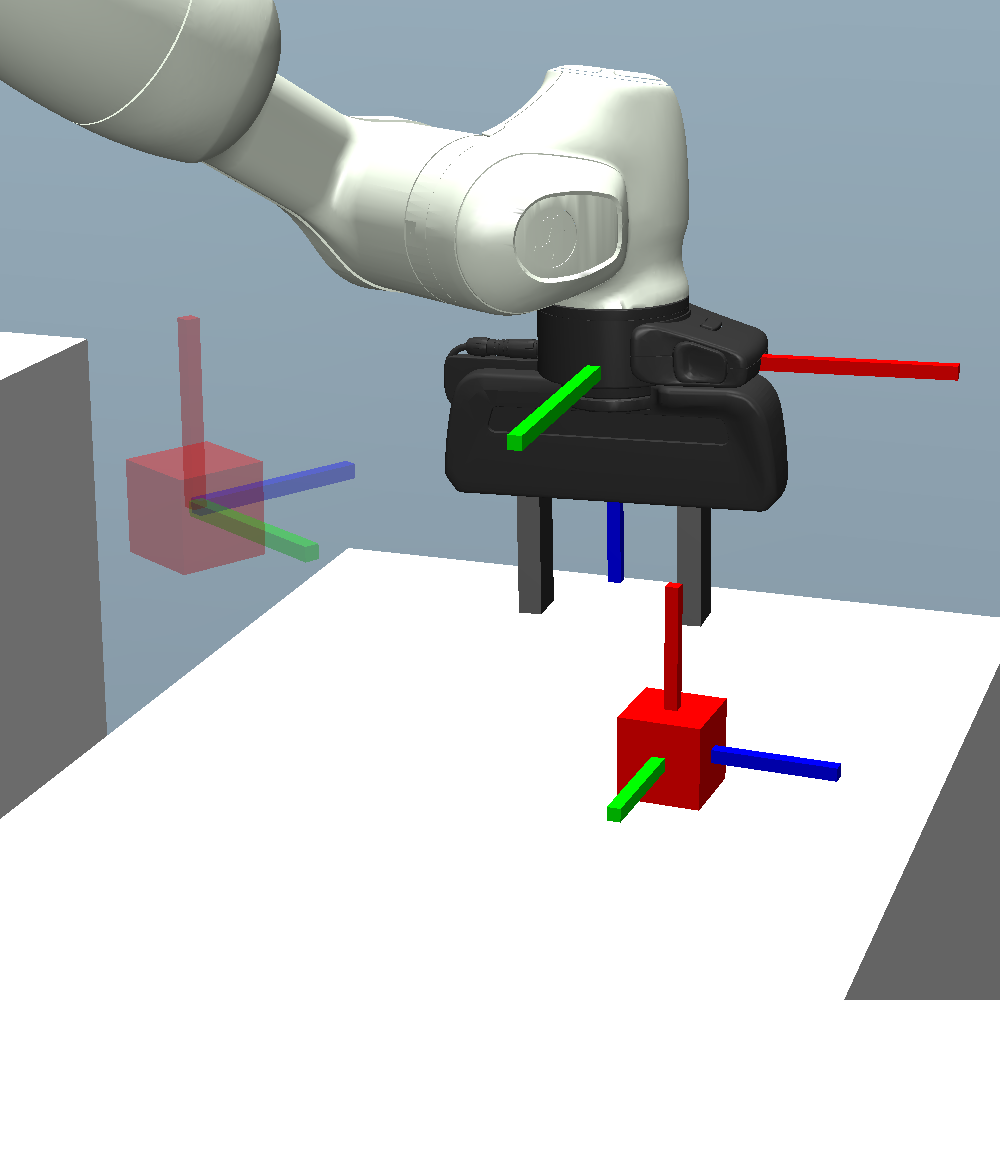}
    \end{subfigure}
    \begin{subfigure}[b]{0.78\linewidth}
        \centering
        \includegraphics[width=\linewidth]{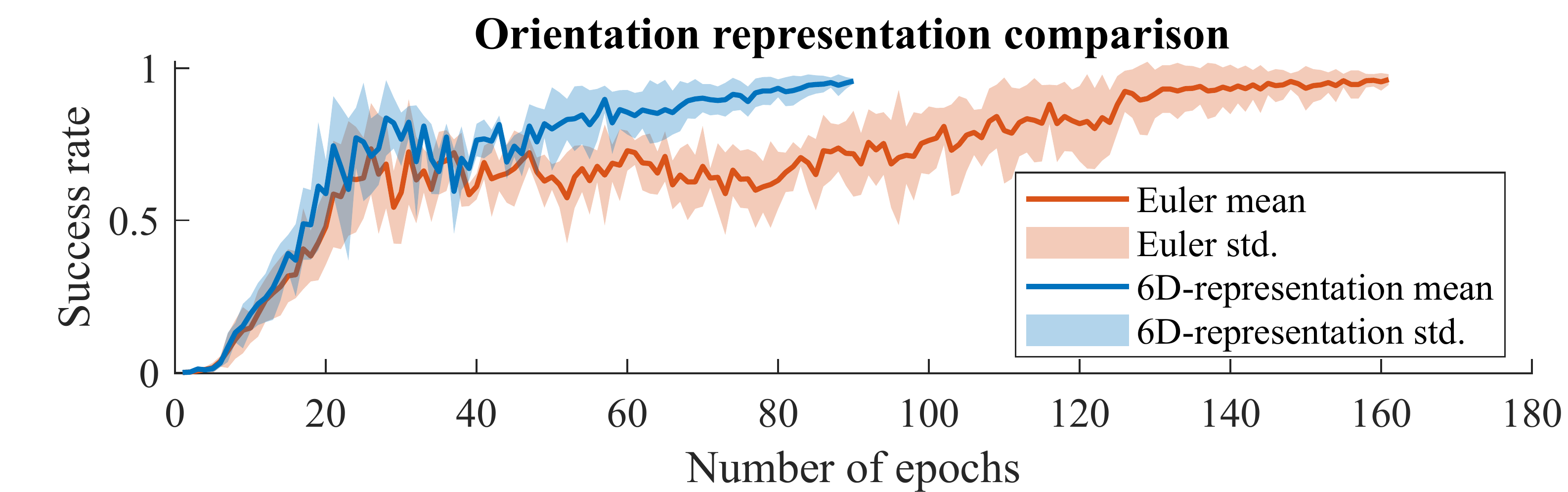}
    \end{subfigure} 
    \caption{Comparison of grasping success rates for a parallel jaw gripper (left) during training. (right) Training with orientation represented as Euler angles in red, training with six-dimensional orientation representation in blue.}
    \label{fig:orientation}
\end{figure}

\paragraph{Full orientation control}
In order to demonstrate the ability of the six-dimensional orientation head to control the full gripper pose and goal orientation, we train a parallel jaw gripper to grasp an object and lift it into a target position and orientation. For comparison, a different agent is trained with an Euler angle representation for the gripper orientation action. The reward function is 0 when the object is within the tolerances of its desired target pose, and -1 else. We sample the goal orientation from a fixed initial orientation that is randomly rotated around the z-axis. Initial learning of grasping is facilitated by starting with an orientation goal tolerance of $\pi$ radians and iteratively reducing the tolerance to a minimum of 0.2 radians when the agent achieves a 0.75 success rate. This iterative approach is motivated by the fact that successful policies for tighter orientation tolerances are a subset of successful policies for larger tolerances. The results for this comparison are displayed in Fig. \ref{fig:orientation}.

Both network architectures learn to place the object in its target position with a 0.75 success rate within approximately 20 epochs. As the orientation tolerance decreases, the performance of both agents drops and the agents adjust their policy to incorporate the orientation goal. Our orientation head reaches the required precision for a 0.95 success rate within 53 and 90 epochs, whereas a network with Euler angles requires between 108 and 161 epochs.

The experiment shows that the choice of action representation indeed affects the ability of the network to learn the optimal policy, and that our orientation head is advantageous to standard discontinuous representations for full pose control. The significance of our finding is underlined by the fact that the representation is more effective despite its higher dimensionality, which is commonly considered disadvantageous for learning.

\begin{figure}[htb!]
    \centering
    \begin{subfigure}[b]{0.2\linewidth}
        \centering
        \includegraphics[width=\linewidth]{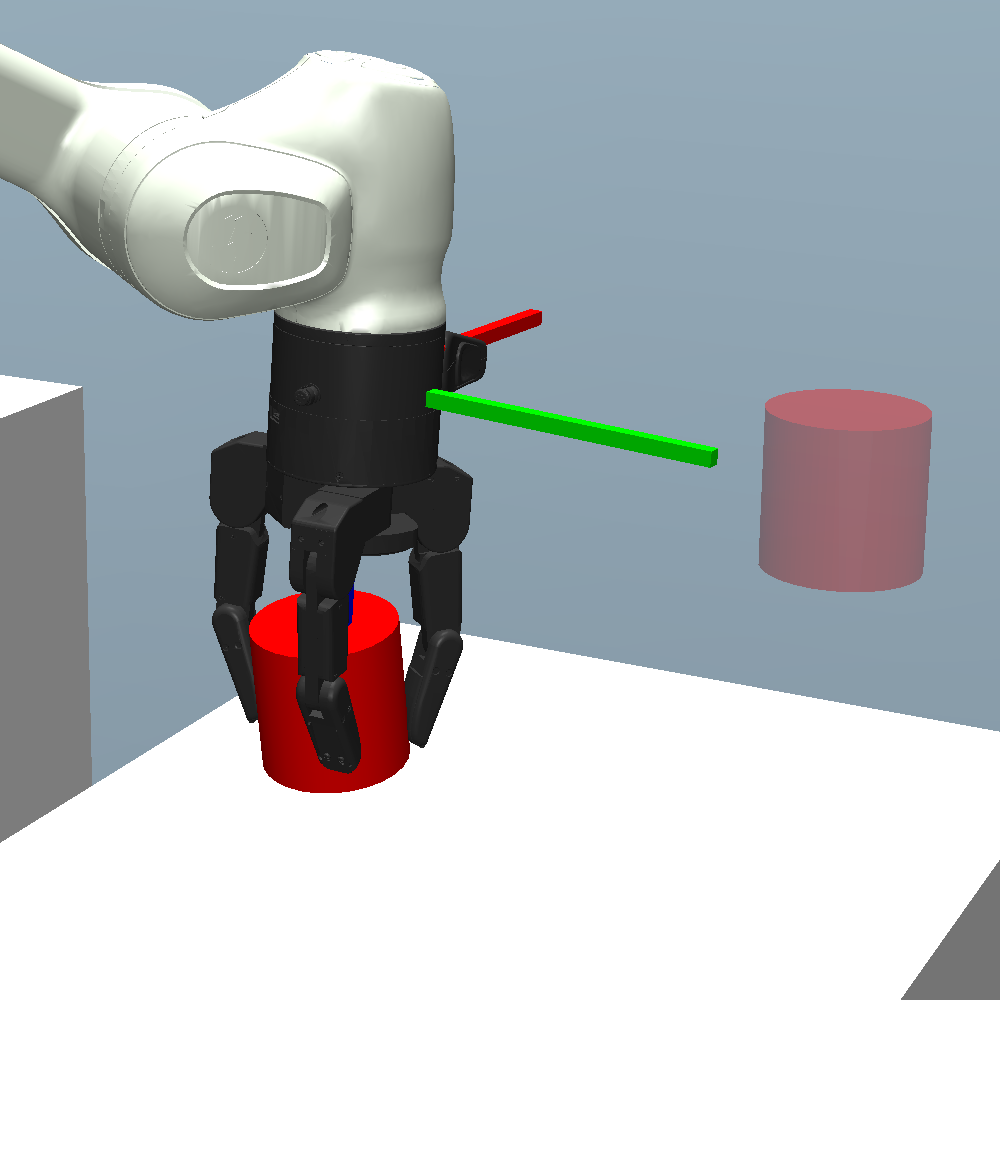}
    \end{subfigure}\hfill
    \begin{subfigure}[b]{0.26\linewidth}
        \centering
        \includegraphics[width=\linewidth]{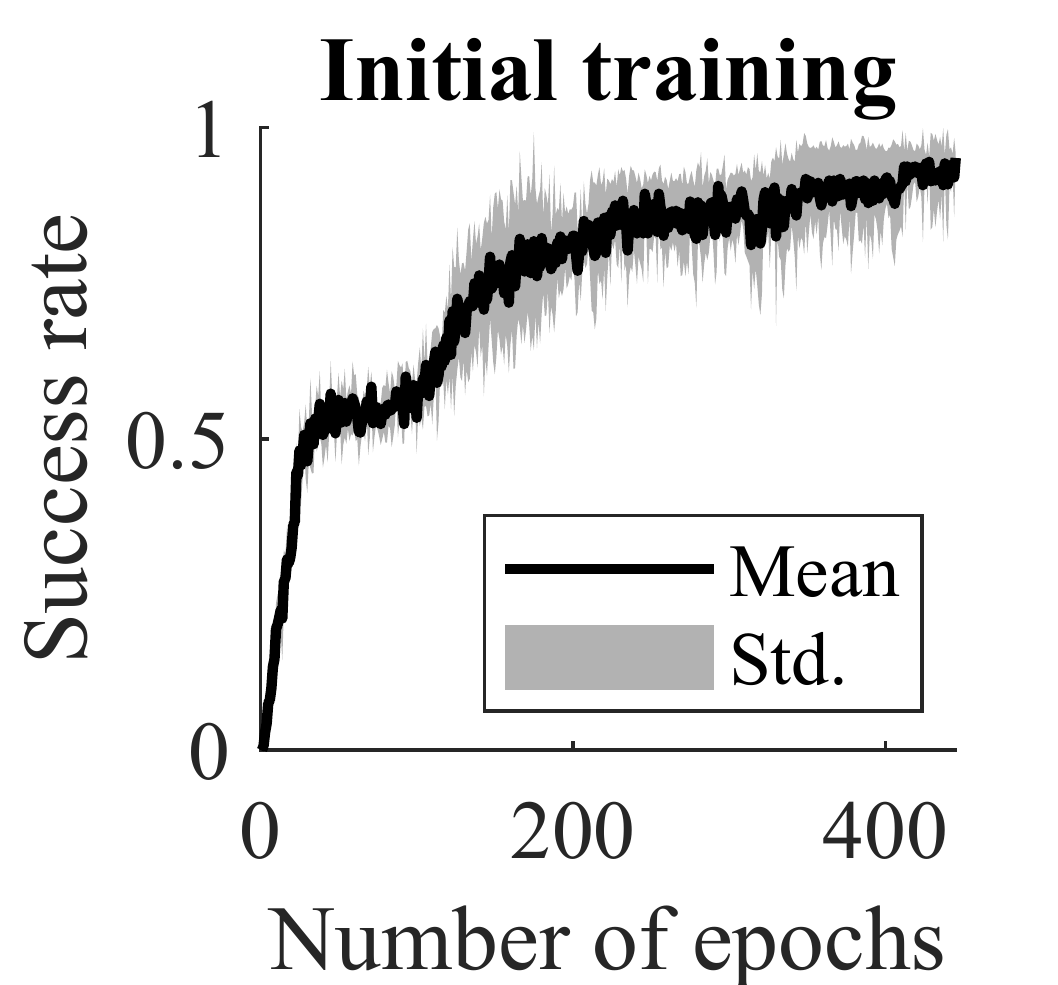}
    \end{subfigure} 
    \begin{subfigure}[b]{0.52\linewidth}
        \centering
        \includegraphics[width=\linewidth]{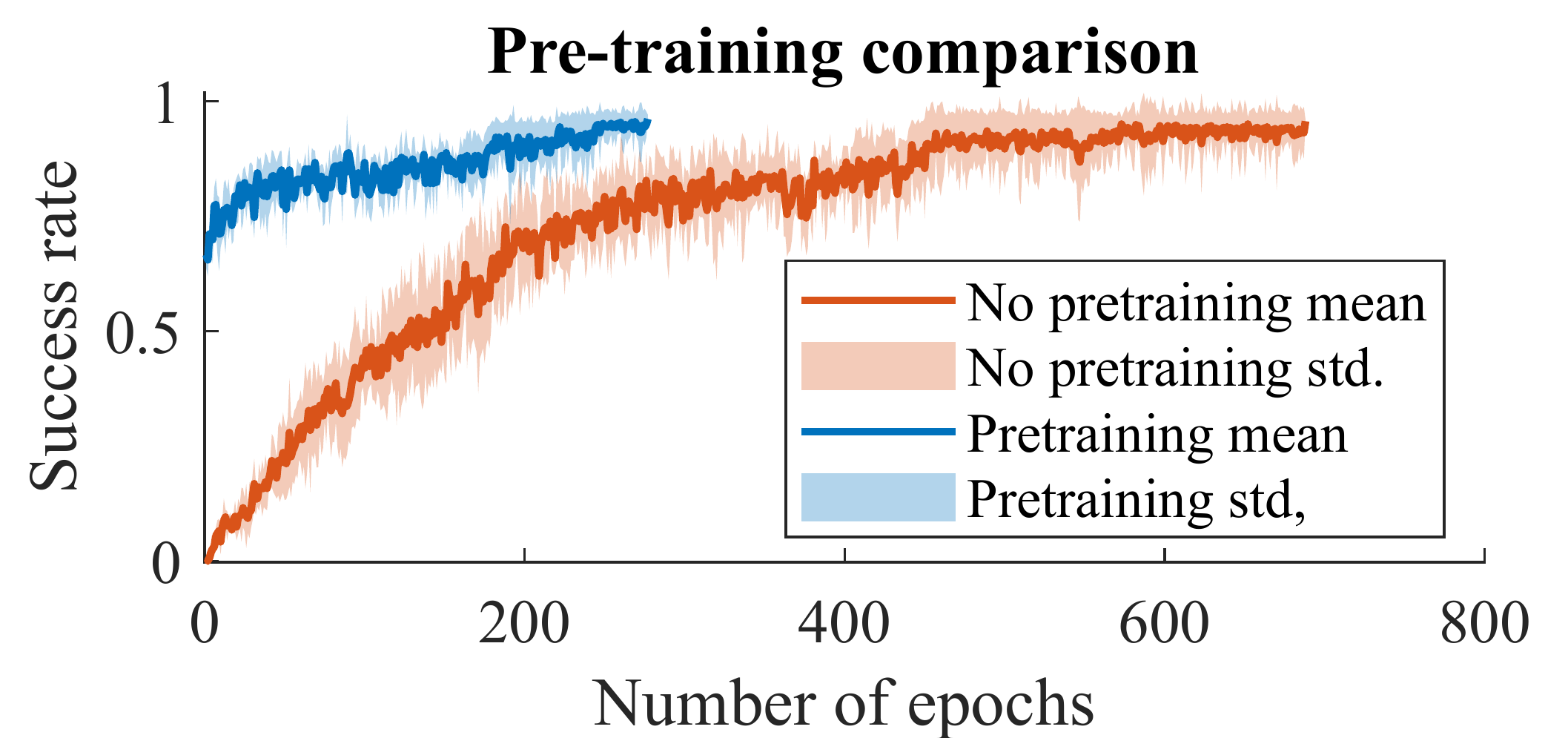}
    \end{subfigure} 
    \caption{Comparison of grasping success for the BarrettHand (left) during training. (center) Initial training on a sphere. (right) Comparison of training on a cube with pre-training in blue and without pre-training in red.}
    \label{fig:pretraining}
\end{figure}

\paragraph{Pre-training}
Training to grasp a specific object requires the agent to learn the general idea of approaching, grasping, and lifting its target object into a goal position repeatedly from scratch every time. In this experiment, we examine the effects of using pre-trained networks as initialization for the actor and critic networks. The initial networks are trained on a sphere of similar dimensions as the cylinder. Figure \ref{fig:pretraining} shows the success rate of a pre-trained agent compared to an agent learning from scratch.

The pre-trained agents start out with a success rate of approximately 0.7 and converge to a policy above 0.95 in between 160 and 277 epochs, whereas the untrained agents achieve the threshold between 407 and 688 epochs. In addition, the standard deviation of the success rate during training of pre-trained agents is smaller than for untrained agents.

The faster convergence of the pre-trained networks to a successful policy indicates that the agent is able to transfer its general knowledge about grasping when trained on a new object. The initial success rate and monotonous performance increase further implies that the underlying universal value function is indeed similar enough to only require fine tuning of the grasp policy. If this was not the case, one would have expected a lower initial success rate and an increased training time as the agent would have to unlearn the previous grasping policy before being able to learn the current task. 
We attribute the reduced variance of the training to the fact that the pre-trained agents do not have to discover the critical idea of how to lift an object by exploration, but only have to apply this concept to a new geometry.

\begin{figure}[hb!]
    \centering
    \begin{subfigure}[b]{0.2\linewidth}
        \centering
        \includegraphics[width=\linewidth]{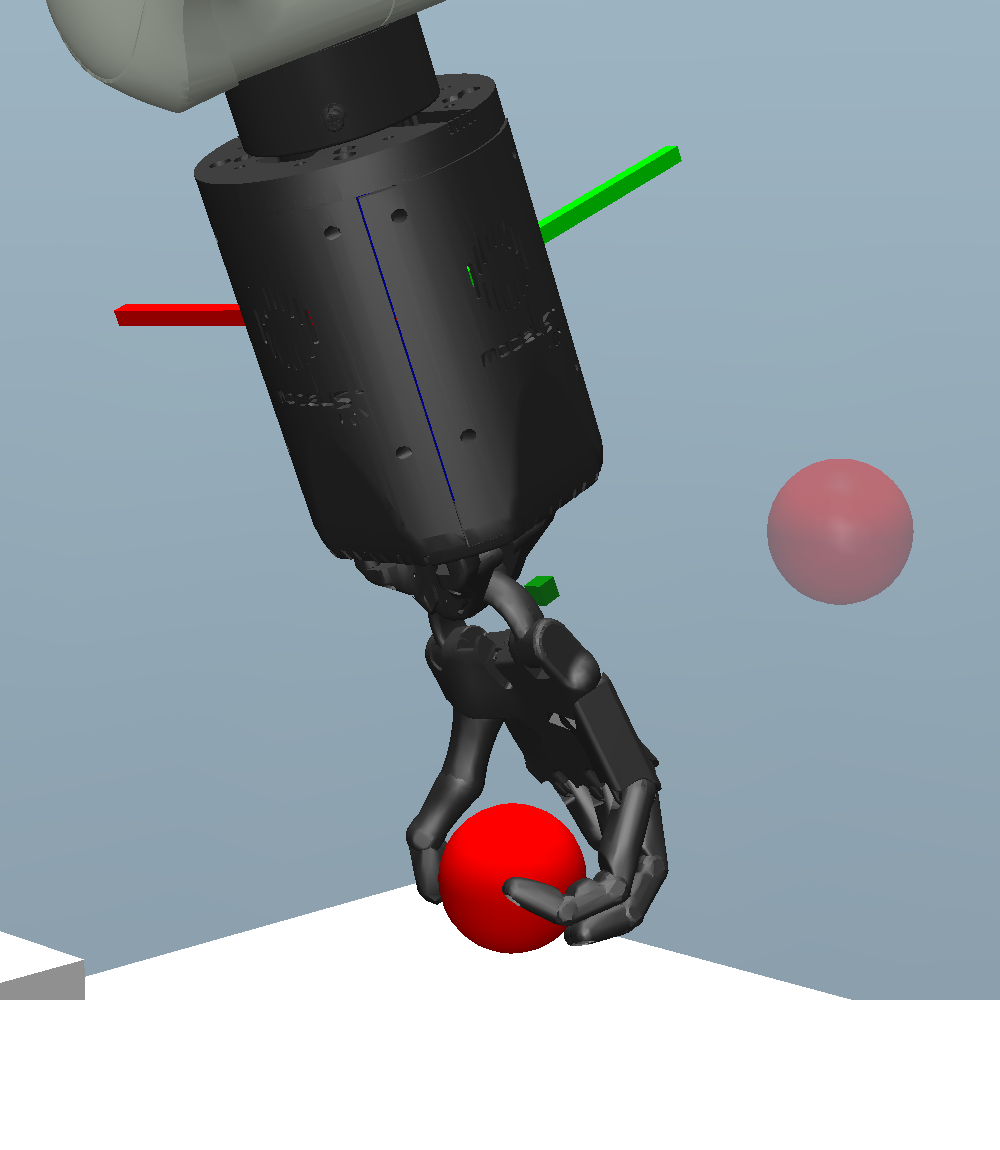}
    \end{subfigure}\hfill
    \begin{subfigure}[b]{0.78\linewidth}
        \centering
        \includegraphics[width=\linewidth]{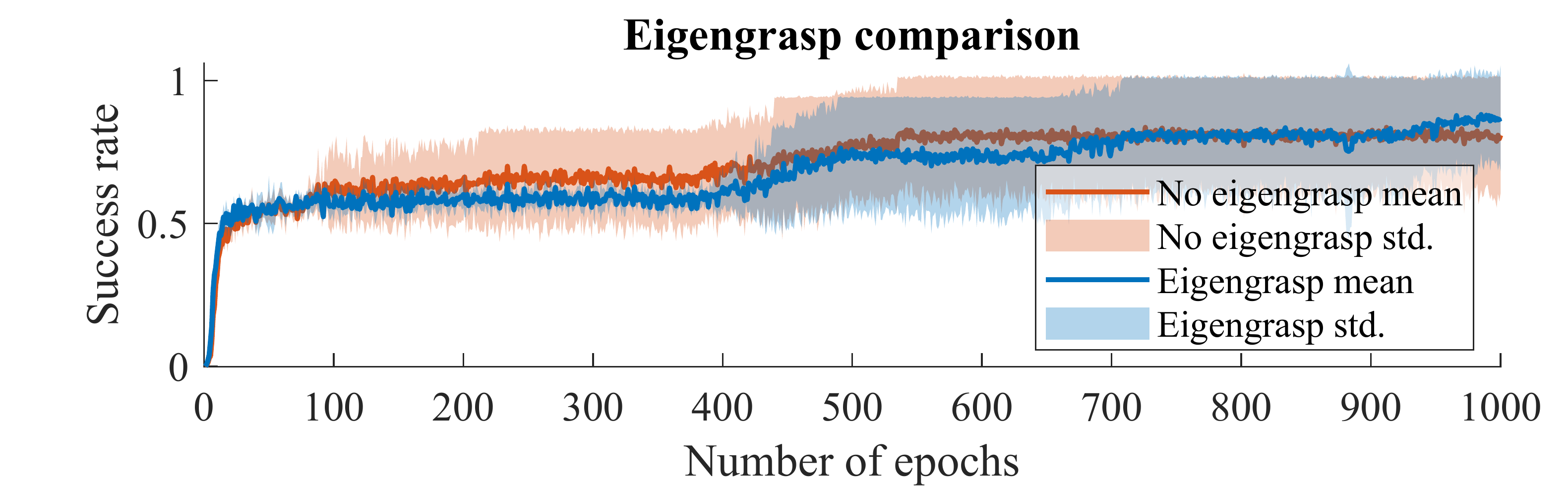}
    \end{subfigure} 
    \caption{Comparison of grasping success for the ShadowHand (left) during training. (right) Training using a state space based on eigengrasps in blue, training on the full state space in red.}
    \label{fig:eigengrasps}
\end{figure}

\paragraph{Eigengrasps}
We validate the inclusion of prior knowledge with the ShadowHand by comparing an agent that controls a superposition of the first seven eigengrasps against an agent with full control over the individual joints. The agents are trained on a sphere as depicted in Fig. \ref{fig:eigengrasps}, with the goal of lifting the sphere into a target position that is randomly sampled on each episode start.

Both agents quickly develop policies that achieve a success rate of around 0.5, i.e, they manage to move objects on the table. Afterwards the agents remain at levels of 0.5 for roughly 400 epochs until they improve to success rates of 0.9 with eigengrasp actions, and 0.8  without eigengrasps. 

We observe that agents develop the critical idea of lifting an object more slowly with complex grippers. Once the concept is understood, the individual agents quickly converge to a successful policy. However, the time frame of necessary learning varies considerably in each run, with one eigengrasp agent and two full control agents not converging within 1000 epochs. This fact also explains the observed plateaus at 0.5 as well as the high variance of the results.

While the success rate with eigengrasps is 0.1 higher compared to no eigengrasps, a definitive statement about the effectiveness of eigengrasps in this particular experiment is difficult due to the observed variance in the training process. We also note that the diversity of our principal components is determined by the human grasp examples, and the quality of our eigengrasps is limited by the dataset's grasp type coverage. ContactPose includes around 2000 useful grasp examples. However, we noticed a lack of precision grasp samples, which directly translates to a less diverse set of actions that our agent can use. Incorporating more grasp data from future datasets should greatly enhance the agent's capabilities to control the hand with low-dimensional actions. At the same time, the results also indicate a detrimental effect of miss-guided prior knowledge.

\begin{figure}[h!]
    \centering
    \begin{subfigure}[b]{0.2\linewidth}
        \centering
        \includegraphics[width=\linewidth]{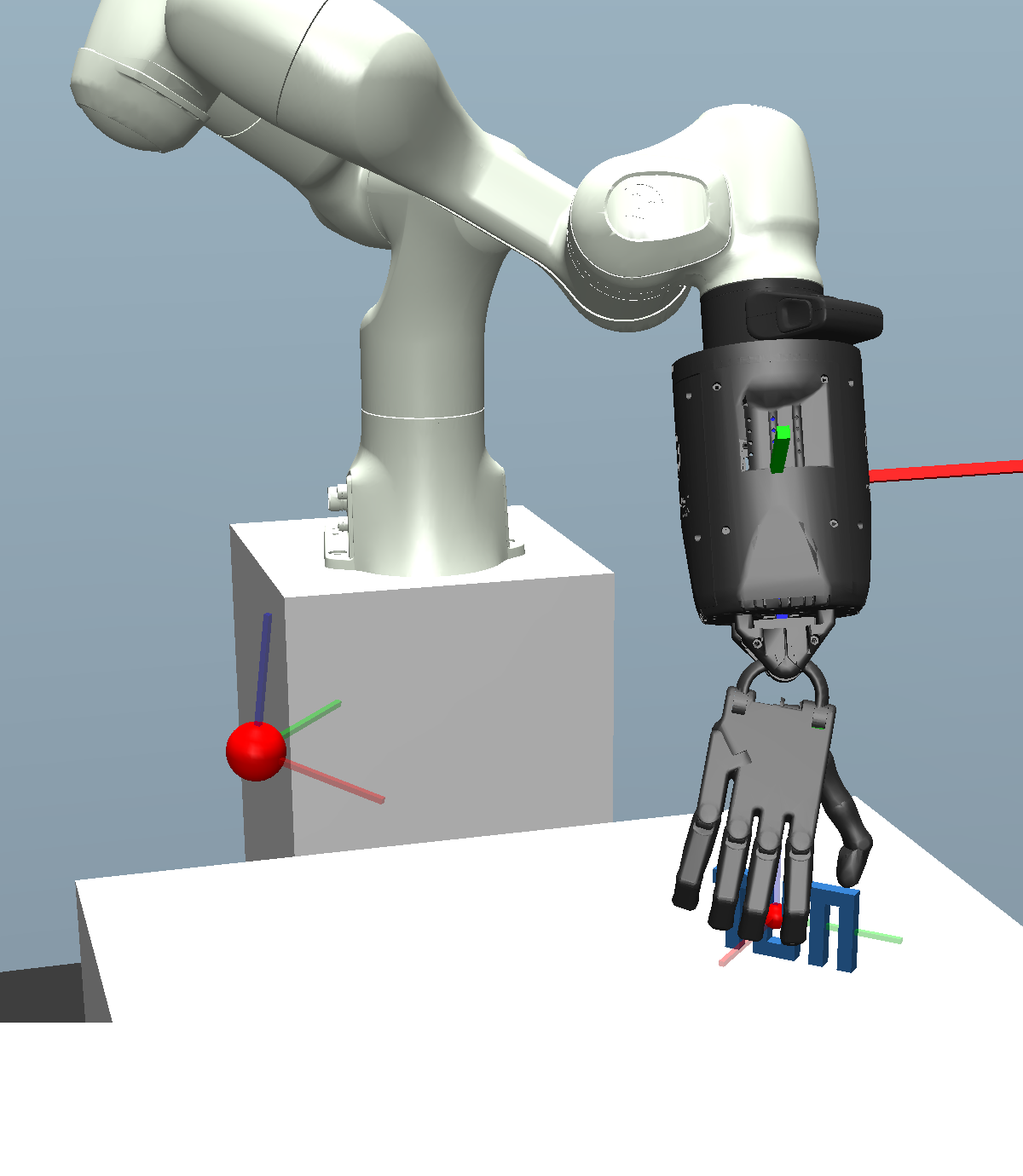}
    \end{subfigure}\hfill
    \begin{subfigure}[b]{0.2\linewidth}
        \centering
        \includegraphics[width=\linewidth]{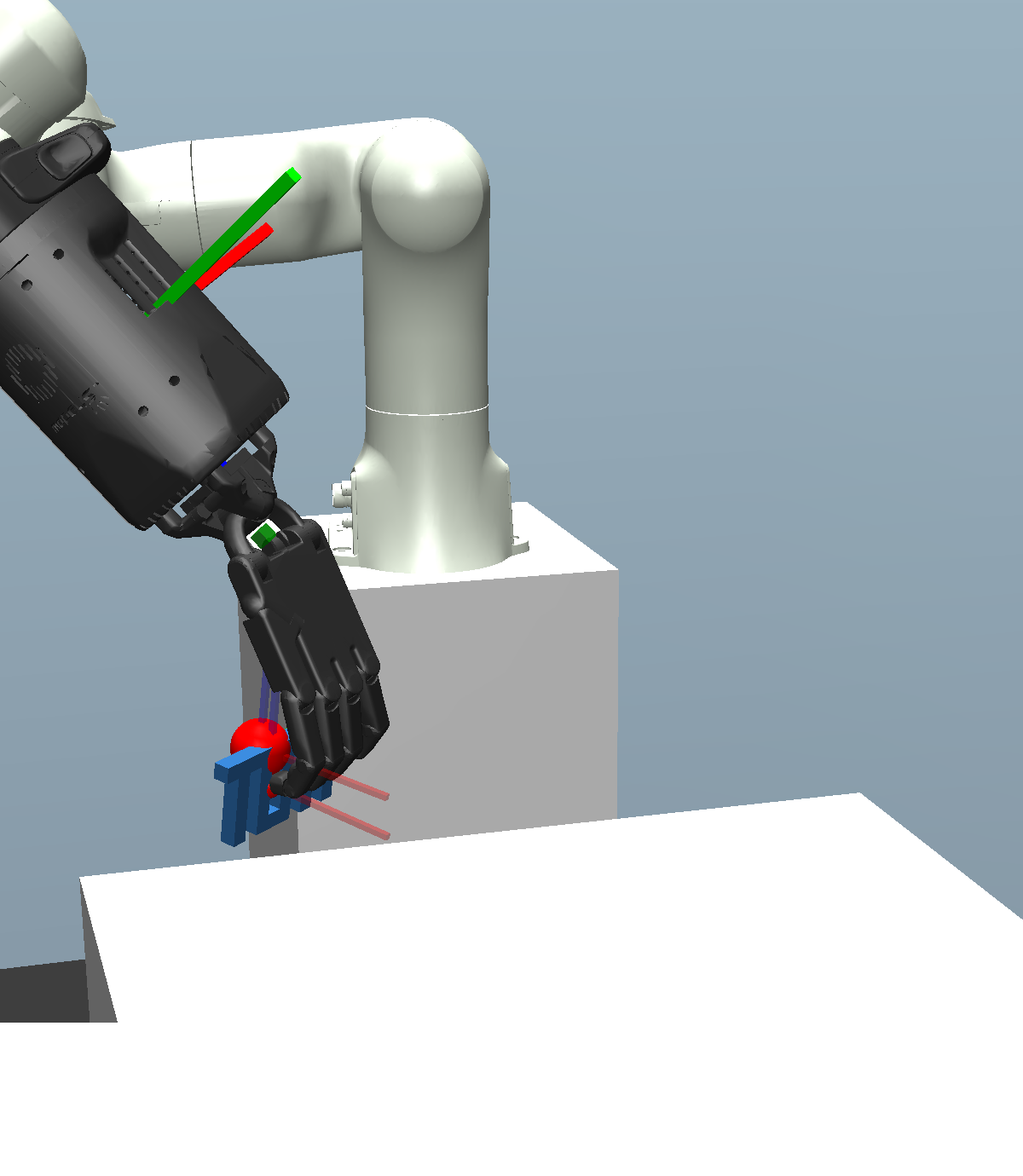}
    \end{subfigure}\hfill
    \begin{subfigure}[b]{0.58\linewidth}
        \centering
        \includegraphics[width=\linewidth]{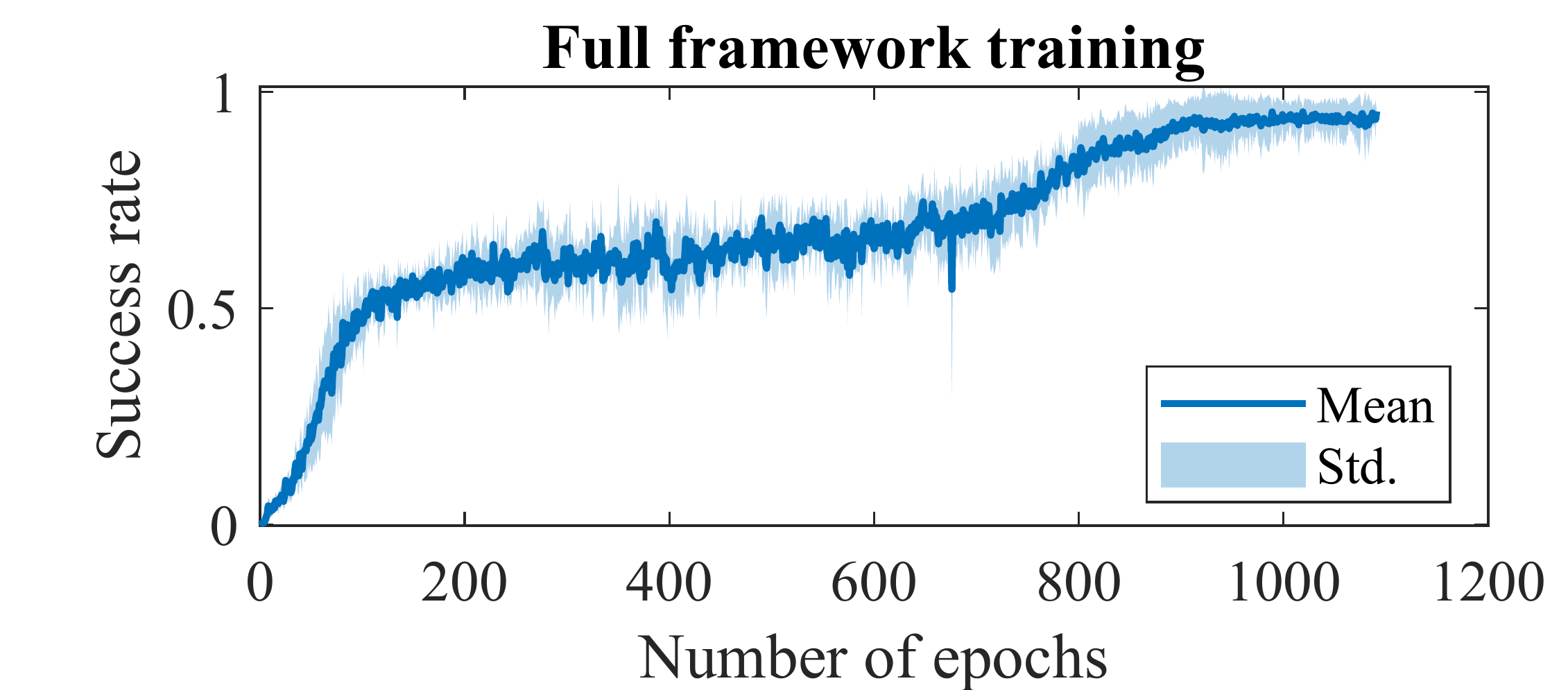}
    \end{subfigure} 
    \caption{Demonstration of grasping a complex shape with the ShadowHand. (left) Object approach. (center) Object at goal position. (right) Training duration.}
    \label{fig:fancy_grasp}
\end{figure}

\paragraph{Full grasping framework}
As a final experiment, we demonstrate the successful combination of the methods presented throughout this paper. The ShadowHand is trained to grasp a complex shape and place it into a randomly sampled target pose. The ShadowHand is controlled using seven eigengrasps. To the best of our knowledge, learning to grasp an object and control its goal pose with a dexterous hand through sparse rewards and without prior expert knowledge has not been demonstrated before. The resulting success rates during training are displayed in Fig. \ref{fig:fancy_grasp}.

During the experiment, the success rate increases to 0.5 within approximately 100 epochs, and then slowly increases to a success rate of 0.95 after around 900 epochs. The variance of the success is significantly reduced compared to the sphere experiment.

Even though the object geometry is more complex, it favors random success in grasping more than a smooth sphere. Accordingly, the variance of the training process is reduced, as agents develop the idea of grasping more consistently. During the gradual increase in performance from a 0.5 to 0.95, the agent learns to both grasp the object and to tune its policy for hand and gripper orientation to achieve full control over the object pose.

\section{Limitations}\label{sec:limitations}
Grasping policies can be learned in simulation using our framework. However, for deployment on real systems, knowledge about the pose of the object that is to be grasped needs to be available. While this information is typically available in a simulated environment, it has to be estimated for real systems, for example through object pose estimation \cite{Xiang2018, Wang2019, Hu2020}. For a parallel jaw gripper, this method was already successfully demonstrated \cite{Andrychowicz2017}.

Our framework does not directly consider an object's geometry in the state space representation. Therefore, an agent is either trained on a single object, or learns one grasp for multiple geometries when trained on different objects without knowledge about the current object's shape. This limitation can be overcome by directly using sensory information, such as point clouds, for the state space representation. In this manner, both the object's pose and shape can be take into account. Learning to grasp from point clouds is possible and has been demonstrated with simpler grippers \cite{mousavian2019, Wang2021}.

A limitation of hindsight learning is that the agent sometimes does not develop the key idea to succeed in its environment. In our sphere experiment, agents sometimes only learn to push objects around to reach goals on the surface. With sufficient time for exploration, they are able to learn the key idea of grasping and lifting the object. However, the time it takes for the agents to develop this crucial understanding varies heavily, which hints at a suboptimal exploration strategy. Potential remedies include the incorporation of intelligent exploration schemes \cite{Houthooft2016} and different noise types, and should be systematically explored in future work.

\section{Conclusion} \label{sec:conclusion}
We have presented a general reinforcement learning framework for grasping with arbitrary grippers and full six-dimensional pose control of gripper and object without requiring prior expert knowledge.

The experiments show that it is possible to train a reinforcement learning agent to grasp even with complex grippers and sparse rewards. A key observation we make is that using networks pre-trained on simple geometric shapes, for example a ball, is extremely effective when learning to grasp new objects. We believe this result arises because universal value function approximators for the grasping problem are largely similar across similar object geometries and only need to be tuned in small areas of their input space. These findings suggest that, if a state representation varies continuously in the object geometry, it should be possible to learn increasingly complex grasps by building upon policies trained on simple geometric objects without any prior knowledge and only using sparse reward functions. As the number of different training objects grows, the agent would then have to start generalizing to unknown objects.

In order to verify the intuition about grasp generalization and to test our approach on real hardware, an object's pose needs to be detected. This detection should be part of the state representation as sensory-based information, such as point clouds, to improve the agent's understanding of the object's geometry and generalize its policy to unknown shapes. Our grasping framework can be extended to support sensory-based input by using a PointNet-like architecture \cite{Qi2017} for the actor and critic.


\clearpage
\acknowledgments{We thank Petar Bevanda, Samuel Tesfazgi, Maximilian Beier, and Armin Lederer for their input and help in preparing this manuscript. This work was supported by the European Union’s Horizon 2020 research and innovation programme under grant agreement no. 871295 ”SeaClear”.}


\bibliography{paper}  

\newpage

\begin{appendices}

    \section{Related work}\label{app:relwork}
    With the advancement of deep learning, the research focus in robotic grasping has largely shifted from analytical methods, for example \cite{Ferrari1992,Goldfeder2007}, to learning-based methods.
    
    One line of research considers grasping as a supervised learning problem. Sample grasps from simulation can be used to create a synthetic dataset for training a convolutional neural network on red-green-blue-depth (RGBD) images and predict grasp success probabilities \cite{mahler2017}. Similarly, voxel-based networks can be used to predict both the best gripper pose and its success probability for each voxel \cite{breyer2020}. Supervised learning is also used to train a variational grasp sampler and a separate grasp quality prediction network that enables the optimization of grasp proposals through the network's derivative \cite{mousavian2019}.
    All of these methods fundamentally differ from our method in their supervised learning principle. Furthermore, they are restricted to parallel jaw grippers, whereas our framework includes dexterous robotic hands.
    Supervised learning for dexterous grasping can be achieved by annotating objects with semantic correspondences to functional areas of the human hand to learn to grasp with an anthropomorphic robotic hand \cite{Zhu2021}. In contrast, we do not require special labeling of each object to map the gripper position to predefined areas.
    In general, supervised learning methods assume that grasp-sampling heuristics used for dataset generation cover all possible grasps. However, this assumption is problematic since it does not hold for most sampling implementations \cite{Eppner2019}. 

    Approaches for grasping based on reinforcement learning are interesting due to their inherent closed-loop nature and nuanced grasping strategies. At the same time, a fundamental problem in applying reinforcement learning to grasping is that grasp success is extremely unlikely with initially random behavior, yet it is necessary for the agent to learn successful policies. 

    A common strategy to overcome this issue is to include prior knowledge about grasping in the reward to increase the success probability, for example by shaping a reward function to promote stable grasps. This principle is used in simulation with object pose information and trust-region policy optimization \cite{Merzic2019}, or with proximal policy optimization \cite{Schulman2017, Wu2019, Shahid2020} to learn grasping from complex reward functions. While reward shaping is helpful to guide training, it introduces a bias in the training towards policies that are effective at maximizing the specific reward function, but not necessarily optimal for accomplishing the desired task \cite{Rajeswaran2018, Trott2019}. As the complexity of a robot's environment grows, it becomes increasingly difficult to find a reasonable reward shape that matches the intended task and provides sufficient guidance during training.
    
    Another idea for facilitating initial learning is introducing off-policy expert knowledge. Grasping and other manipulation tasks can then be trained with natural policy gradients augmented with a penalty term for deviations from human demonstrations previously recorded in virtual reality with motion capture \cite{Rajeswaran2018}. Similarly, an expert dataset can be generated from standard motion planning and grasp planning tools to penalize policies that differ from the expert dataset and thereby guide the agent's exploration \cite{Wang2021}. Along the same line of reasoning, teleoperation trajectories can be recorded to improve the initial grasping guess with path integral policy improvement \cite{montforte2021}. A drawback of these approaches is that expert knowledge must be obtained for each task and imposes a data-driven prior on the training process. This prior introduces a bias towards policies that are close to the demonstrations, but might not be optimal for the agent's task.
    
    Instead of relying on external insight into complex problems, hindsight learning improves the sample efficiency of sparse reward experience. The intended goal is swapped with an achieved goal during training to obtain samples with a positive reward \cite{Andrychowicz2017, Plappert2018}. While this concept is evaluated in simulation and the transfer to real systems is demonstrated, the algorithms and examples in \cite{Andrychowicz2017} related to object relocation are limited to a parallel jaw gripper with fixed orientation and disregard the object's goal orientation. 
    
    Sparse binary grasp-indicating rewards are sometimes replaced with or supplemented by grasp-quality-metric-based rewards to promote robust grasps \cite{montforte2021, koenig2022}. These works build upon a rich literature in grasp metrics based on the notion of force closure and the grasp wrench space \cite{Bicchi1995, Ferrari1992}.

    Training robotic grasping with reinforcement learning on real hardware is still uncommon, with only a few notable exceptions \cite{Levine2018, Kalashnikov2018}. A more widely used strategy is to train in simulation, and later deploy the trained models with minor adaptions on real hardware \cite{Wu2019, Wang2021}. The same principle has also been applied to policies trained with hindsight learning with pose estimators and additional noise during training \cite{Andrychowicz2017}.

    \section{Reinforcement learning environments} \label{app:environmets}

    Each episode consists of 50 time steps over the course of two seconds, where the agent receives an observation and chooses its action. Internally, a time step is further divided into 20 simulator steps of 0.002 seconds that continuously apply the last action chosen by the agent. The agent's observation consists of the gripper pose, the gripper joint angles, the gripper velocity, the target object pose, the target object pose relative to the gripper pose, and the target object's velocity. Each goal is given as a goal position in world coordinates, or a position and orientation given as six-dimensional representation described in Appendix \ref{app:orient} for environments with orientation goals. All orientation states are given in the six-dimensional orientation representation. The agent's action consist of the desired translation of the gripper link, its orientation in the six-dimensional representation, and its gripper joint positions.
    The gripper joint positions are normalized to an interval of $[-1, 1]$, where -1 is the lower joint limit, and 1 is the upper joint limit. If eigengrasps are used, each gripper action corresponds to the weight of the eigengrasp in the superposition of joint angles.

    On episode start, a random object orientation and goal position are sampled, and the arm is placed into a random position facing down onto the table. Half of the goals are sampled on the surface of the table, and half are sampled above the table. Our agent receives a reward of $0$ while the object's distance to its goal is smaller than the distance threshold of 0.05, and a reward of -1 else. For environments with additional target orientation, the reward is 0 when the object is within the distance threshold of 0.05, and its orientation deviation is below the angular distance threshold. In orientation environments, we decrease the orientation threshold from an initial value of $\pi$ to 0.2 radians. The threshold is reduced by a factor of 1.25 whenever the agent has reached a 0.75 success rate.

    All environments are simulated in MuJoCo. Each gripper is mounted to the seven DoF Franka Panda robot arm. The agent only controls the gripper position and orientation with MoJoCo's \textit{mocap} mechanics. The arm configuration is calculated by MuJoCo to satisfy the commanded gripper pose while taking the arm constraints into account. The grippers as well as all test objects use four-dimensional contact and friction models.
    
    The parallel jaw gripper has a only one degree of freedom and symmetrically closes its jaws. We use a position actuator in MuJoCo with $kp=30000$. The BarrettHand has four degrees of freedom. It can rotate two of its fingers around the spread joint of the palm link, where the two fingers are coupled at a 1:1 ratio, and rotate the proximal joint of each finger. The distal joint on each finger is coupled with the proximal joint at a 3:1 ratio. While the real BarrettHand's joint model also includes a more complex \textit{breakaway} mechanism, we deem this model sufficient to simulate the kinematics. The BarrettHand uses four position actuators with $kp=200$.
    The ShadowHand features 20 degrees of freedom, with two actuated wrist joints, five actuated thumb joints, 3 actuated finger joints each for the first finger, middle finger and ring finger, and 4 joints for the little finger. Each of the fingers except the thumb features an additional coupled joint for the finger tip at a 705:805 ratio. The wrist tendons are position actuated with $kp=5$. The finger tendons are position actuated with $kp=1$.

    The cube in the orientation head comparison environment has an edge length of 5cm and weighs 2kg. The sphere and the cylinder in the BarrettHand environments both have a radius of 4cm, and the cylinder has a length of 8cm. Both weigh 0.5kg. The sphere in the ShadowHand environment has a radius of 0.35cm and weighs 0.2kg. The complex geometry weighs 0.16kg.

    \section{Training algorithm and hyperparameters} \label{app:training}

    All experiments are conducted using the same architecture and set of hyperparameters. Our training largely follows the algorithm outlined in \cite{Andrychowicz2017}. The actor network consists of four hidden layers with 256 nodes, ReLU activation functions and tanh output activation functions to ensure the actions lie in [-1, 1].
    It uses the action heads described in Sec. 2. The orientation head converts the 6 dimensional orientation output to a full rotation matrix. If the six-dimensional orientation representation consists of two parallel vectors, the conversion fails. With properly trained orientations, this is not an issue. If, however, the tanh functions saturate, parallel vectors become a distinct possibility. To prevent this failure mode, we further add a regularization loss to the actor's loss function that penalizes saturated outputs. The critic network also uses four hidden layers with 256 nodes with ReLU activations, but without output activation.
    We clip the objective to the interval of $[-200, 200]$ to avoid the unbounded divergence of value estimates (see \cite{Hasselt2018}).
    To improve stability during training, we further employ target networks with polyak averaging updates for both the actor and the critic. Our replay buffer samples 80\ of its experience with hindsight goals observed in the same episode after the sampled experience.
    The exploration noise consists of actions chosen uniformly at random with a probability of $\epsilon$, and additive Gaussian noise for the actions chosen by the actor. Each experiment runs on 16 nodes according to the distributed approach outlined in Sec. 2.
    A full list of hyperparameters is given in Table \ref{tab:hyperparameters}.

    \begin{table}[h]
    \begin{center} 
        \begin{tabular}{|l c|}
        \hline
        Hyperparameter & Value\\ [0.5ex] 
        \hline
        actor learning rate & 0.001\\
        \hline
        critic learning rate & 0.001\\
        \hline
        $\tau$ & 0.05 \\ 
        \hline
        $\epsilon$ & 0.3\\
        \hline
        $\mu$ & 0.0\\
        \hline
        $\sigma$ & 0.2\\
        \hline
        $\gamma$ & 0.98\\ [1ex] 
        \hline
        maximum epochs & 1000\\
        \hline
        cycles & 50\\
        \hline
        rollouts & 2\\
        \hline
        training epochs & 40\\
        \hline
        batch size & 256\\
        \hline
        HER ratio $k$& 4\\
        \hline
        replay buffer size & 1,000,000\\
        \hline
        early stopping success rate & 0.95\\
        \hline
        \end{tabular}
        \caption{Training hyperparameters.}\label{tab:hyperparameters}
    \end{center}
\end{table}

    In addition to the shared networks, all nodes also share a normalizer to estimate the mean and standard deviation for each input and normalize all input values except the orientations to zero mean and unit variance.
    Our hindsight experience replay buffer uses the "future" strategy outlined in \cite{Andrychowicz2017}.
    The full algorithm is listed in Algorithm \ref{alg:DDPG}. We stop the training if the agent achieves a success rate of over 0.95, or if the maximum number of epochs has been reached. All experiments are repeated five times with random seeds 1, 50, 100, 150, and 200.

    \begin{algorithm}
        \caption{DDPG algorithm}
        \label{alg:DDPG}
        \begin{algorithmic}[1] 
            \State Initialize critic network $Q_\phi(s, a)$ and actor network $\pi_\theta(s)$ with random or pre-trained weights
            \State Broadcast network weights from node 0 to all other nodes
            \Statex On all nodes do:
            \State Initialize target networks $Q'$ and $\pi'$ with weights $\phi' \leftarrow \phi$, $\theta' \leftarrow \theta$
            \State Initialize replay buffer B
            \For{episode = 1:N\_episodes}
                \For{cycle = 1:N\_cycles}
                    \For{rollout = 1:N\_rollouts}
                        \State Observe initial state $s_{1}$, goal $g$ and achieved goal $ag_1$
                        \For{t = 1:T}
                            \If{$rand() \leq \epsilon$}
                                \State Select action $a_t$ uniformly at random
                            \Else
                                \State Select action $a_t = \pi_\theta(s) + \mathcal{N}_t$, where $\mathcal{N}_t$ is the exploration noise
                            \EndIf
                            \State Take action $a_t$ and observe new state $s_{t+1}$ and achieved goal $ag_t$
                            \State Store transition $(s_t, a_t, ag_t, s_{t+1}, ag_{t+1})$ in B
                            \EndFor
                        \State Update normalizers with all samples and synchronize
                        \For{epoch = 1:N\_epochs}
                            \State Sample minibatch of $N$ transitions $(s_t, a_t, s_{t+1}, g)$ from B
                            \State Calculate rewards $r_t$ with new hindsight goals
                            \State Set $y_i = r_i + \gamma Q'_{\phi'}(s_{i+1}, \pi_\theta(s_t))$
                            \State Compute critic gradient by minimizing $L = \frac{1}{N} \sum_i(y_i - Q_\theta(s_i, a_i))^2$
                            \State Compute actor policy gradient as
                            \Statex \begin{align*}\nabla_\theta J \approx \frac{1}{N}\sum_i \nabla_a Q_\theta(s, a)|_{s=s_i, a=\pi_\theta(s_i)} \nabla_\theta \pi_\theta(s)|_{s=s_i}\end{align*}
                            \State Synchronize actor and critic gradients and apply the update step
                        \EndFor
                        \State Update the target networks locally as
                        \Statex \begin{align*}\phi' \leftarrow \tau \phi + (1 - \tau) \phi' \\ \theta' \leftarrow \tau \theta + (1- \tau) \theta' \end{align*}        
                    \EndFor
                \EndFor
            \EndFor
        \end{algorithmic}
    \end{algorithm}

    \section*{The orientation action head} \label{app:orient}
    The choice of input and output representation has a major impact for the performance of neural networks. In particular, it has been shown that the smoothness of a function is a key factor in the accuracy of the trained approximation. Meanwhile, orientations have many equivalent representations that are used in robotics, such as Euler angles, axis angles, quaternions and rotation matrices. A critical observation of \cite{Zhou2019} is that most of these representations are a discontinuous representation of $SO(3)$, the space of orientations, in the sense that a continuous one-to-one mapping from $SO(3)$ to the representation space $R$ exists. Consequently, even if the underlying universal value function and optimal policy approximated by the neural networks vary smoothly in the object's orientation, the optimal actions can be discontinuous if an unsuitable representation is chosen.

    An easy way to see this is to consider a gripper pose close to a discontinuity in its Euler representation. If the object pose and the corresponding optimal gripper alignment vary slightly in their orientation, the corresponding optimal orientation action would be discontinuous, which is difficult to learn. Conversely, an action representation as presented in \cite{Zhou2019} guarantees that target orientations close to each other are also close in the network's representation.

    The steps to recover a full orientation matrix from the six-dimensional representation are described in Algorithm \ref{alg:embedding}. This procedure can fail in three cases, namely if either of the vectors $a_1$ and $a_2$ is zero, or if the vectors are parallel. We omit the first two cases since it is extremely unlikely for a network to output an exact zero vector. The same argument applies to the parallel case, however, as the action head uses a tanh activation function and these are known to suffer from saturation of values, the networks sometimes converge to saturated regions of the activation function, where the output is either -1 or 1. For these cases, parallel vectors are a distinct possibility. We therefore add a regularization term for the orientation action to the training loss. Regularization does not bias towards specific orientations as the vector's magnitude has no impact on the orientation.

    \begin{algorithm}
        \caption{Orientation recovery. Adapted from \cite{Zhou2019}.}
        \label{alg:embedding}
        \begin{algorithmic}[1] 
            \State Extract the six-dimensional representation vector $x$ from the actor network and subdivide it into two vectors $a_1$ and $a_2$.
            \State $b_1 \gets \frac{a_1}{||a_1||}$
            \State $b_2 \gets a_2 - (b_1 \cdot a_2)b_1$
            \State $b_2 \gets \frac{b_2}{||b_2||}$
            \State $b_3 \gets b_1 \times b_2$
            \State \Return $[b_1, b_2, b_3]$
        \end{algorithmic}
    \end{algorithm}

    \section*{Eigengrasps for the ShadowHand} \label{app:eigengrasps}
    In order to extract eigengrasps by principal component analysis, it is necessary to have a diverse dataset of hand configurations. Since, to the best of our knowledge, there exists no dataset of successful grasps for the ShadowHand, we base our analysis on the ContactPose dataset \cite{Brahmbhatt2020}. The dataset contains the joint poses of human hands from over 2000 human grasp trials. We assume that due to the similar kinematics of a human hand and the ShadowHand, a successful human grasp is also a valid grasp choice for the ShadowHand.

    The hand configurations from the ContactPose dataset are visually examined and outliers with likely bad sensor readings removed. Subsequently, the corresponding anthropomorphic robotic hand joint configuration is obtained by minimizing the distance between each human hand link and its robotic counterpart with an inverse kinematics solver. In order to be able to achieve the best fit to the hand pose, the grasp pose is rescaled to the size of the ShadowHand. Using the dataset of joint angles from imitated grasps, we can now extract the hand synergies by running a principal component analysis. A demonstration of the procedure with the ShadowHand and the final eigengrasps are available at our eigengrasp repository.

\begin{figure}
        \centering
        \begin{subfigure}[b]{0.45\linewidth}
            \centering
            \includegraphics[width=\linewidth]{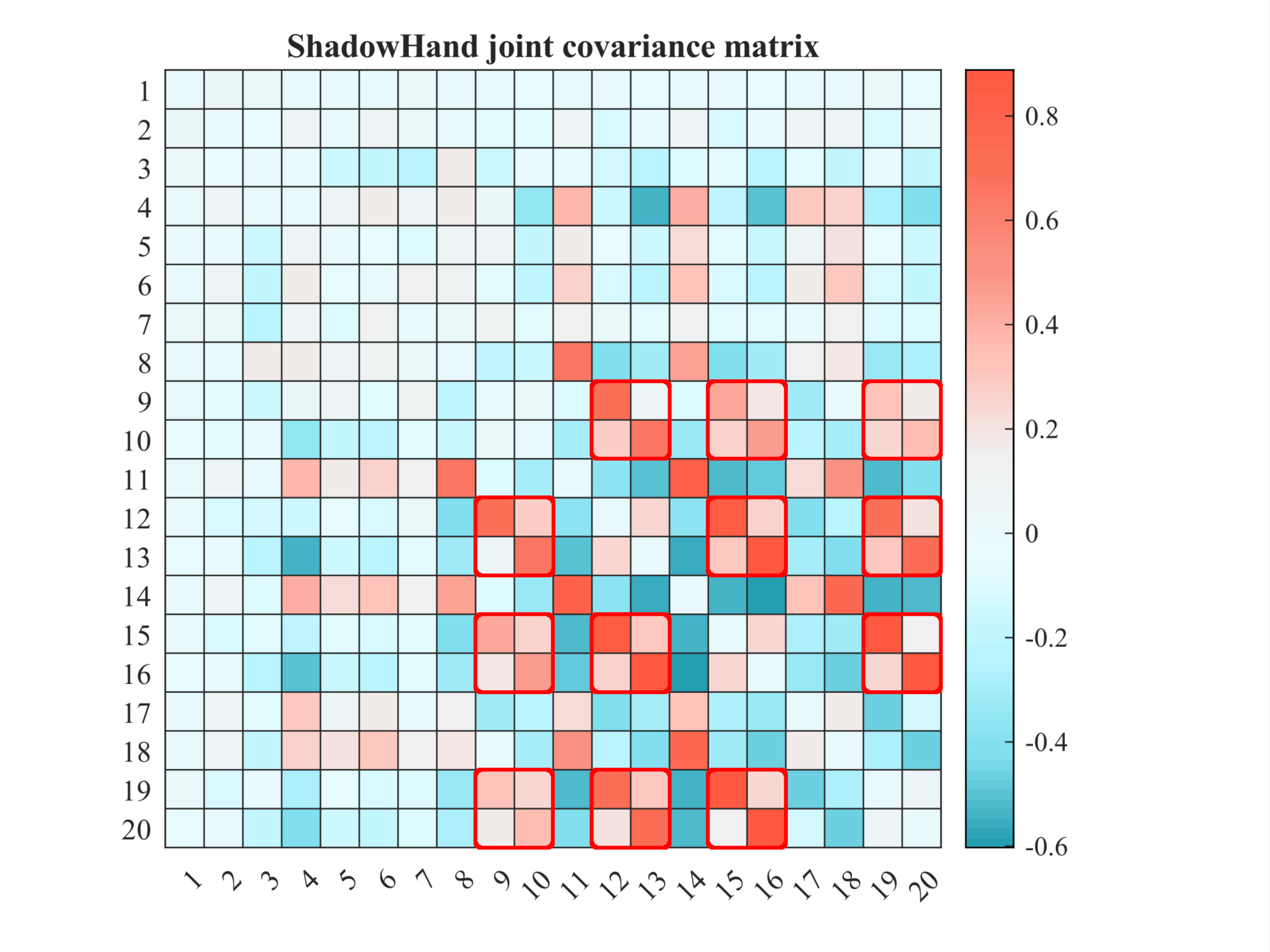}
        \end{subfigure}\hfill
        \begin{subfigure}[b]{0.45\linewidth}
            \centering
            \includegraphics[width=\linewidth]{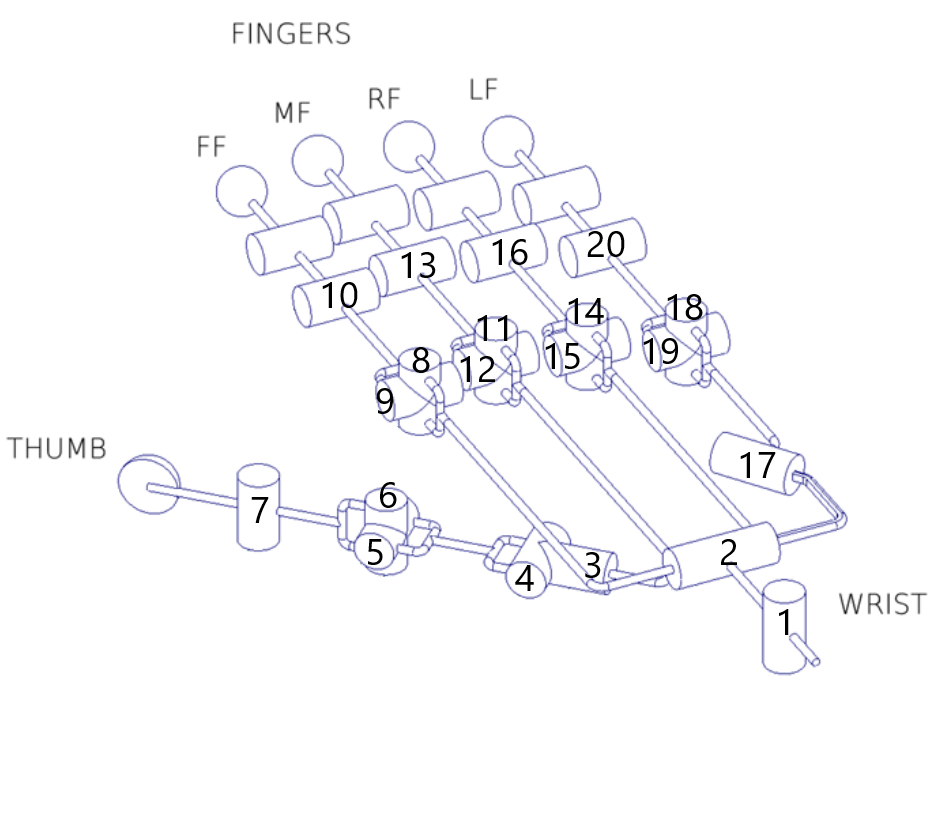} 
        \end{subfigure} 
        \caption{Analysis of the eigengrasp data obtained for the ShadowHand. (left) Covariance matrix of the grasp pose data. A clear structure in the correlation of the joint space is visible. Diagonal entries are zeroed out to emphasize off-diagonal elements. The palm block structure is highlighted for clarity. (right) ShadowHand schematic with marked joints. Coupled fingertip joints are not included in the eigengrasps. Picture adapted from \cite{shadowhand}.}
        \label{fig:eigen_combined}
    \end{figure}

    An analysis of the covariance matrix of joint data shows that the dataset's joint configurations are indeed highly structured. The heatmap of the matrix in Fig. \ref{fig:eigen_combined} reveals a block-wise correlation for the palm joints responsible for a power grasp (joint number 9, 10, 12, 13, 15, 16, 19 and 20), as well as some weaker correlations between the finger's lateral motion joints and the thumb joints. This reaffirms the belief that much of a the grasp shape's variance can be explained in a lower-dimensional subspace of the joint configuration space. To help with the identification of joints in the covariance matrix, Fig. \ref{fig:eigen_combined} depicts a schematic drawing of the ShadowHand with joints marked by their respective number. 
    
\begin{figure}
        \centering
        \includegraphics[width=0.6\linewidth]{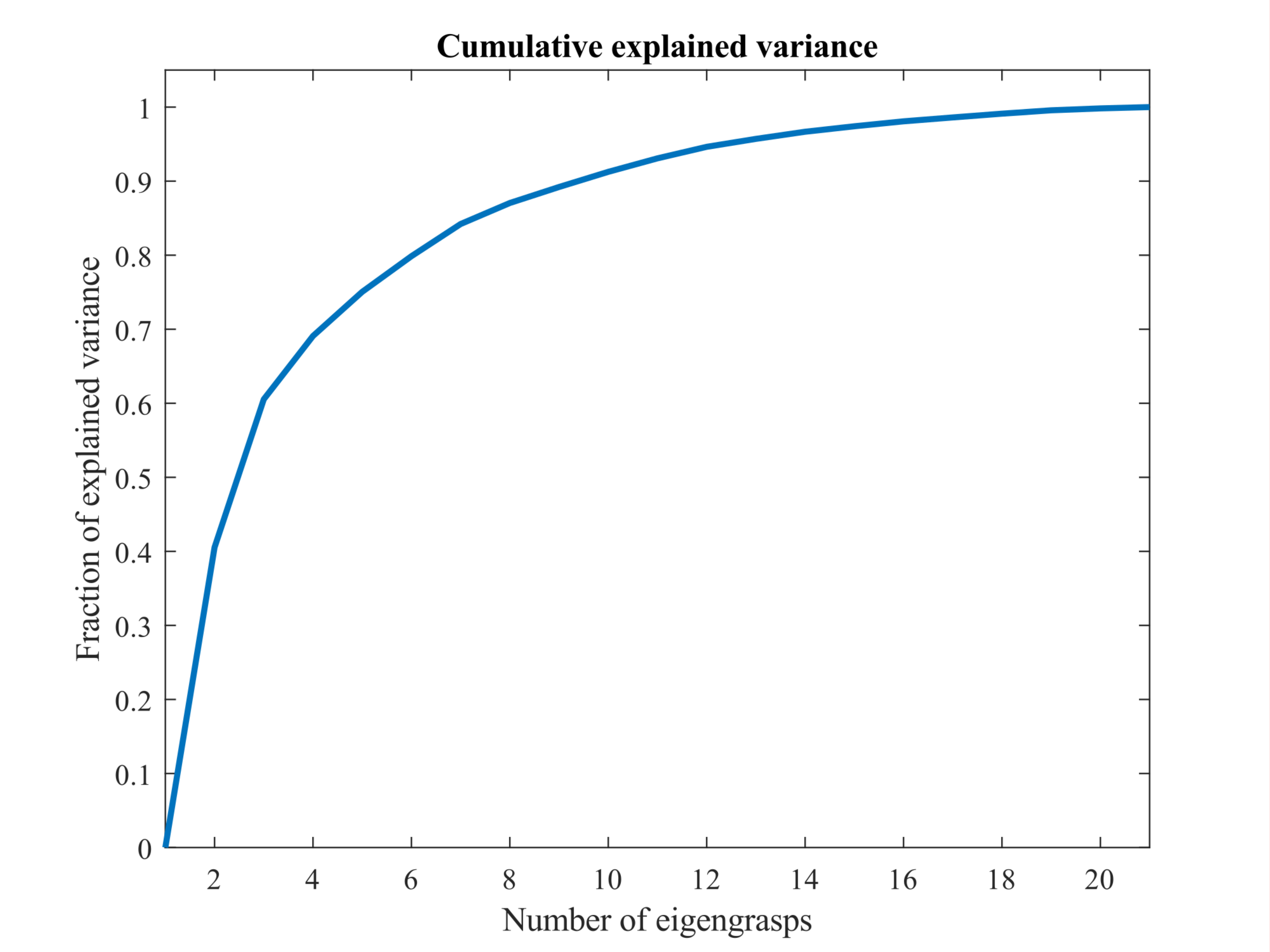}
        \caption{Cumulative fraction of data variance explained by the first N eigengrasps. The first synergy axes contain the bulk of information, while the last axes yield diminishing increases.}
        \label{fig:eigen_pca}    
    \end{figure}

    The cumulative fraction of data variance explained by the first N singular values is depicted in Fig. \ref{fig:eigen_pca}. An increasing number of eigengrasps yields diminishing returns, so that a classic tradeoff between precision and accuracy based on the variance of the grasps is possible.

    One limitation of this approach is that the frequency of specific grasp types and the diversity of the grasp dataset both have a large influence on the extracted components. Further datasets with human grasp poses would greatly aide in refining these eigengrasps. Alternatively, synergies can also be designed by hand.
\end{appendices}

\end{document}

%% file: acronyms.tex
\begin{acronym}
    \acro{MDP}[MDP]{Markov decision process}
    \acro{DDPG}[DDPG]{deep deterministic policy gradient}
\end{acronym}